\documentclass[a4paper,twoside]{article}
\usepackage{makecell}
\usepackage{epsfig}
\usepackage{subcaption}
\usepackage{calc}
\usepackage{amssymb}
\usepackage{amstext}
\usepackage{amsmath}
\usepackage{amsthm}
\usepackage{multicol}
\usepackage{pslatex}
\usepackage{apalike}
\usepackage{algorithm2e}
\usepackage{dblfloatfix}
\usepackage{xcolor}
\usepackage{bm}
\usepackage{siunitx}
\usepackage{nomencl}
\makenomenclature
\usepackage[bottom]{footmisc}
\usepackage[draft,inline,nomargin,index]{fixme}
\fxsetup{theme=color,mode=multiuser}
\definecolor{adamnote}{rgb}{0.349,0.804,0.565}
\definecolor{zdeneknote}{rgb}{0.247, 0.655, 0.839}
\FXRegisterAuthor{ad}{ak}{\color{adamnote}Adam}
\FXRegisterAuthor{zd}{zh}{\color{zdeneknote}Zdenek}
\DeclareMathOperator*{\argmax}{arg\,max}

\newcommand\given[1][]{\:#1\vert\:}
\usepackage{SCITEPRESS}     
\begin{document}
\title{Parameter Adjustments in POMDP-Based Trajectory Planning for Unsignalized Intersections}

\author{\authorname{Adam Kollarčík\sup{1,2}\orcidAuthor{0000-0002-9150-839X}, and Zdeněk Hanzálek\sup{2}\orcidAuthor{0000-0002-8135-1296}}
\affiliation{\sup{1}Dept. of Control Engineering, Faculty of Electrical Engineering, Czech Technical University in Prague, Czech Republic}
\affiliation{\sup{2}Czech Institute of Informatics, Robotics and Cybernetics, Czech Technical University in Prague, Czech Republic}
\email{\{adam.kollarcik, zdenek.hanzalek\}@cvut.cz}
}

\keywords{Automated Intersection Crossing, Trajectory Planning, POMDP}

\abstract{This paper investigates the problem of trajectory planning for autonomous vehicles at unsignalized intersections, specifically focusing on scenarios where the vehicle lacks the right of way and yet must cross safely. To address this issue, we have employed a method based on the Partially Observable Markov Decision Processes (POMDPs) framework designed for planning under uncertainty. The method utilizes the Adaptive Belief Tree (ABT) algorithm as an approximate solver for the POMDPs. We outline the POMDP formulation, beginning with discretizing the intersection's topology. Additionally, we present a dynamics model for the prediction of the evolving states of vehicles, such as their position and velocity. Using an observation model, we also describe the connection of those states with the imperfect (noisy) available measurements. Our results confirmed that the method is able to plan collision-free trajectories in a series of simulations utilizing real-world traffic data from aerial footage of two distinct intersections. Furthermore, we studied the impact of parameter adjustments of the ABT algorithm on the method's performance. This provides guidance in determining reasonable parameter settings, which is valuable for future method applications.}


\onecolumn \maketitle \normalsize \setcounter{footnote}{0} \vfill

\section{\uppercase{Introduction}}
\begin{figure}[!h]
  \centering
{\epsfig{file = 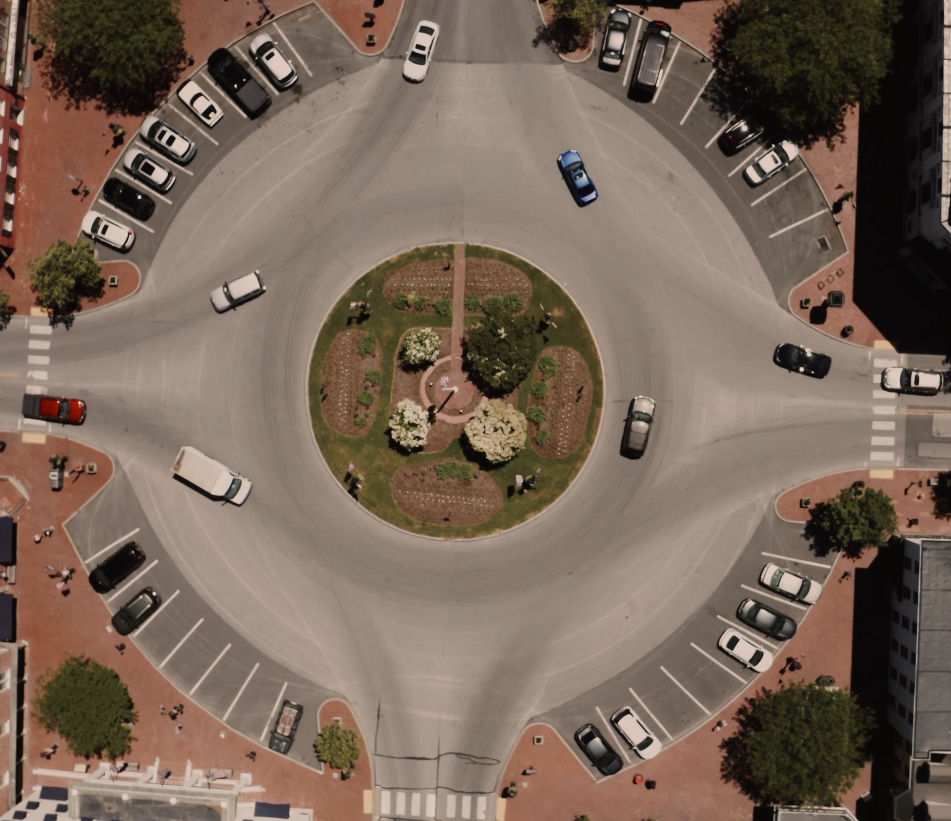, width = 0.9\columnwidth}}
  \caption{Unsignalized intersection (roundabout).}
  \label{fig:roundabout_photo}
 \end{figure}

The advent of autonomous vehicles signals a change in urban mobility, promising to enhance efficiency, safety, and the overall driving experience. One of the challenges these vehicles face is crossing an unsignalized intersection, e.g., a roundabout shown in Figure~\ref{fig:roundabout_photo}. With the absence of traffic signals and especially without the right of way, the autonomous vehicle must independently determine whether and how to proceed, guided only by traffic regulations and the inherently uncertain behavior of other vehicles, perceived by measurements affected by noise. This is the description of the trajectory planning problem for unsignalized intersections we explore in this paper.

In particular, we investigate a method based on Partially Observable Markov Decision
Processes (POMDPs), a framework for planning under uncertainty with incomplete information. For a given intersection and 2D measurements of other vehicles, the method outputs the trajectory as a sequence of accelerations along the desired path.

Our contributions are twofold. At first, we verify
the capability of the POMDP-based method to plan collision-free trajectories in simulations using real-life traffic data from aerial footage of two intersections with different topologies\footnote{See: video link unavailable for double-blind review}. In addition, based on those simulations, we analyze the influence of the parameter values on the method's performance.\label{introduction}
\section{\uppercase{Related Work}}
\label{sec:related_work}
The literature \cite{eskandarianResearchAdvancesChallenges2021a,hubmannAutomatedDrivingUncertain2018a} broadly divides the planning strategies for automated vehicles into rule-based, reactive, and interactive methods, each with distinct approaches to predicting future events. 

The rule-based methods have a predefined set of rules and actions established \cite{schwartingPlanningDecisionMakingAutonomous2018a}, e.g., in the form of a finite state machine. Thus their decisions rely only on the current state.  
The reactive methods treat the motion of vehicles as independent entities.
They may introduce simplifying assumptions such as the constant velocity of other vehicles or their set future behavior \cite{hubmannAutomatedDrivingUncertain2018a}. Together with the rule-based methods, they lack the ability to consider the interconnected behavior of all vehicles, which is essential for safe interactions in automated driving~\cite{schwartingPlanningDecisionMakingAutonomous2018a}.

Interactive methods can be either centralized or decentralized \cite{eskandarianResearchAdvancesChallenges2021a}. The centralized methods are usually based on the Vehicle-to-Everything (V2X) paradigm \cite{tongArtificialIntelligenceVehicletoEverything2019a} and assume communication among vehicles to generate a single global strategy. Nevertheless, given predictions that only half of the vehicles will be autonomous by 2060 \cite{litmanAutonomousVehicleImplementation2023a}, decentralized approaches are vital for the foreseeable future. Decentralized methods might be further categorized into three main groups according to their way of tackling interactivity. Those are game theory-based, probabilistic, and data-driven approaches \cite{schwartingPlanningDecisionMakingAutonomous2018a}. 
 
 This paper investigates a probabilistic planning method developed for unsignalized intersection crossing introduced in \cite{hubmannAutomatedDrivingUncertain2018a} based on POMDPs. The method relies on the Adaptive Belief Tree algorithm \cite{kurniawatiOnlinePOMDPSolver2016}, using a particle filter to estimate the probability distribution of non-observable variables, such as the intentions of other vehicles. In \cite{beyPOMDPPlanningRoundabouts2021a}, the authors employed the method for driving at roundabouts. In \cite{hubmannPOMDPManeuverPlanner2019}, occlusion caused both by the environment and by other dynamic objects was included, and in \cite{schulzLearningInteractionAwareProbabilistic2019a}, an improved behavior prediction with dynamic Bayesian network was introduced.
\section{\uppercase{Approach}}
This section outlines the key components of our approach to addressing the unsignalized intersection crossing problem. We begin by providing a formal problem definition in Section~\ref{subsec::formal_def}. Key to our strategy is formulating this problem as a POMDP, which we discuss in Section \ref{subsec:pomdp}. To solve the POMDP, we apply an approximate solver called the Adaptive Belief Tree (ABT) algorithm, detailed in Section ~\ref{subsec:ABT}. Our method involves discretizing the intersection topology, a process explained in Section~\ref{subsec:topology}, where we identify all feasible 2D paths without lane changes based on combinations of intersection entrances and exits. Subsequently,
we introduce dynamical and observation models in Sections~\ref{subsec:models},~and ~\ref{subsec:obs}, respectively, that capture the complexities
of the real world in a simplified yet sufficiently accurate manner for trajectory planning. Essential entities such as {\it state} and {\it observation} are defined, and we construct a reward function in Section~\ref{subsec:reward}. This function is designed to be maximized within the POMDP, ensuring that the resulting trajectory aligns with our expectations, e.g., encompasses crucial factors such as collision avoidance and adherence to velocity limits. To complete the process, we leverage the already mentioned ABT solver for our POMDP formulation to derive the desired trajectory as a sequence of longitudinal accelerations along the desired path toward the goal exit of the intersection.
\subsection{Formal Problem Definition}
\label{subsec::formal_def}
To formally define the problem, we seek a collision-free trajectory through a known intersection topology along a predefined path  $\bm{\phi}(t): t \rightarrow \mathbb{R}^2,\,\, t\in [0,\ d]$ where $d$ is the total length of the path. We assume that the position $\bm{p}_i = [x_i,\ y_i]^\top$, velocity $\bm{v}_i = [v_{x,i},\ v_{y,i}]^\top$, unit heading vector $\bm{\theta_i}$, width $w_i$, and length $l_i$ measurements of all $n$ relevant other (non-ego) vehicles $i \in \{1, \dots, n\}$ are available every sample time $k$. Thus, we are looking for a sequence of longitudinal accelerations $a_0^k$ of the ego vehicle (denoted with the zero subscript), such that there is no overlap between the bounding rectangles given by $\bm{p}_0^k$,  $\bm{\theta}_0^k$, $w_0$, $l_0$ and $\bm{p}_i^k$,  $\bm{\theta}_i^k$, $w_i$, $l_i$ for each $i$. Lateral accelerations of all vehicles need not be directly determined, as the curvature of the followed path implicitly controls it, provided that drifting is avoided.

\nomenclature{\(\phi\)}{A path}
\nomenclature{\(t\)}{Parametrization parameter of a path}
\nomenclature{\(d\)}{A path lenght}
\nomenclature{\(n\)}{Number of road users (vehicles)}
\nomenclature{\(\bm{p}\)}{Global position}
\nomenclature{\(\bm{v}\)}{Global velocity}
\nomenclature{\(\bm{\theta}\)}{Global unit orientation vector}
\nomenclature{\(k\)}{Control period (time)}
\nomenclature{\(w\)}{Bounding rectangle width}
\nomenclature{\(l\)}{Bounding rectangle width}

\subsection{POMDPs}
\label{subsec:pomdp}
POMDPs generalize the Markov Decision Process (MDP) framework for decision-making and planning under uncertainty \cite{kurniawatiPartiallyObservableMarkov2022}.
They are defined by a tuple $\left< S, A, O, T, Z, R, \gamma\right> $, where $S$ is the set of states, $A$ is a set of actions, $O$ is the set of observations, $T$ is a set of conditional transition probabilities between states, $Z$ is a set of conditional observation probabilities, $R: S \times A$ is the reward function, and $\gamma \in \left(0,\ 1 \right]$ is the discount factor. 

Fundamentally, $T$ represents the stochastic dynamics model of the system, and $Z$ represents the stochastic observation (measurement) model with the probabilities $T(s'|a,s)$ of transition from state $s \in S $ to state $s'\in S$ with action $a\in A $, and $Z(o|a,s')$ of observation $o\in O$ in $s'$. The particular way of modeling the dynamical and measurement properties for our purposes is described Sections \ref{subsec:models} (dynamical model) and \ref{subsec:obs} (observation model).

Since the current state is unknown, the so-called {\it belief} $b \in B$ is introduced, representing a distribution over all states from the set $B$ known as the {\it belief~space}:
\begin{equation}
\label{eq:belief_up}
b(s') \propto Z(o\given[]a,s')\sum_{s \in S} T(s' \given[] a,s) b(s) ,\
\end{equation}
where the $\propto$ sign denotes that the right-hand side probabilities are not normalized.
This converts the original POMDP problem into an MDP where the beliefs are the states, with the continuous fully observable state space B.
\nomenclature{\(S\)}{ POMDP -- Set of states}
\nomenclature{\(A\)}{ POMDP -- Set of actions}
\nomenclature{\(O\)}{ POMDP -- Set of observations}
\nomenclature{\(T\)}{ POMDP -- Set of conditional transition probs.}
\nomenclature{\(Z\)}{ POMDP -- Set of conditional observ. probs.}
\nomenclature{\(R\)}{ POMDP -- Reward function}
\nomenclature{\(a\)}{ POMDP -- action $\triangleq$ acceleration}
\nomenclature{\(o\)}{ POMDP -- observation}
\nomenclature{\(s\)}{ POMDP -- state}
\nomenclature{\(s'\)}{ POMDP -- next state}
\nomenclature{\(B\)}{ POMDP -- Set of beliefs aka. belief space}
\nomenclature{\(b\)}{ POMDP -- belief}

The goal of the POMDP planner is to determine a policy $\pi: B\rightarrow A$, that should maximize the expected  sum of discounted rewards called the {\it value function}~$V$:
\begin{equation}
\label{eq:val_func}
V = \mathbb{E}\left[\sum_{k=0}^{\infty}\gamma^k R(\bm{s}_k,a_k)\given[\Big] b,\pi \right].
\end{equation}
Even though the belief space is continuous, exact POMDP solvers exist.
They are based on the fact that any optimal value function is piecewise linear and convex and thus can be described as an upper envelope of a finite set of linear functions called the  $\alpha$-{\it vectors} \cite{braziunasPOMDPSolutionMethods2003}. 
Nevertheless, exact methods are highly intractable and not suitable for online applications \cite{kurniawatiPartiallyObservableMarkov2022}. Approximate online methods mostly perform a belief tree search combining heuristics, branch and bound pruning, and Monte Carlo sampling \cite{yeDESPOTOnlinePOMDP2017}. 
An example of these approximate methods is the Adaptive Belief Tree algorithm (ABT) implemented with its variants in a publicly available solver called TAPIR \cite{klimenkoTAPIRSoftwareToolkit2014a}, which we use in our work.
\nomenclature{\(V\)}{ POMDP -- value function}
\nomenclature{\(\pi\)}{ POMDP -- policy}
\subsection{Adaptive Belief Tree Algorithm}
\label{subsec:ABT}
The ABT algorithm \cite{kurniawatiOnlinePOMDPSolver2016} is an online and anytime POMDP solver that uses the Augmented Belief Tree for finding well-performing policies. 
The tree $\mathcal{T}$ is constructed by sampling initial belief $b_0$ with $n_\mathrm{par}$ particles and propagating them with actions according to the observation and dynamics models $T$ and $Z$ until the set depth of the optimization horizon $N$ or a terminal state is reached. 
At each step, the particles are also assigned a reward $r= R(s,a)$. 

\nomenclature{\(\mathcal{T}\)}{ POMDP -- belief tree}
\nomenclature{\(r\)}{ POMDP -- reward}
\nomenclature{\(N\)}{ POMDP -- depth of optimization horizon}

\begin{figure}[!h]
  \centering
   {\epsfig{file = 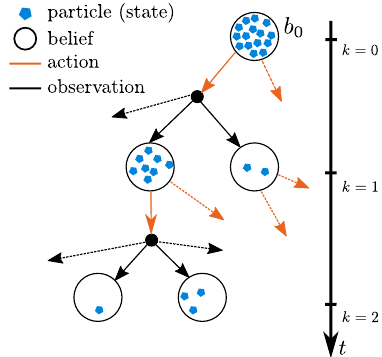, width = 5.5cm}}
  \caption{Illustrative example of an augmented belief tree.}
  \label{fig:tree}
 \end{figure}

Sequences of quadruples $(s_i,a_i,o_i,r_i)$ representing a path in the tree with $i$ being the depth of each contained node are sampled state trajectories also referred to {\it episodes}. 
At last, each episode is assigned an additional quadruple $(s_N,-,-,r_N)$ where $s_N$ is a sampled next state and $r_N$ is the expected future reward for all following states. 
This is computed directly for the terminal states; non-terminal states are heuristically estimated, in our case, with a 3-step lookahead. 
An example of an Augmented Belief Tree is illustrated in Figure \ref{fig:tree}.
 
With the tree containing $n_\mathrm{ep}$ episodes, the policy based on a {\it Q-value} estimate $\hat{Q}(b,a)$  for each belief-action pair is obtained:
\begin{equation}
V(b,\pi) = \max_{a\in{\mathcal{T}(b)}}{\hat{Q}(b,a)},\
\end{equation}
\begin{equation}
\label{eq:policy}
\pi(b) = \argmax_{a\in{\mathcal{T}(b)}}{\hat{Q}(b,a)} ,\
\end{equation}
where Q-value $Q(b,a)$ is a reward for performing an action $a$ in the first step and then behaving optimally. 
The $\hat{Q}(b,a)$ estimate is computed as an average reward of all episodes $h \in H(b,a)$ containing the pair $(b,a)$ in depth~$l$:
\begin{equation}
\hat{Q}(b,a) = \frac{1}{|H(b,a)|}\sum_{h\in H(b,a)}\left(\sum_{i = l}^N{\gamma^{i-l}r_i}\right).
\end{equation}
After the action from $\pi(b)$ is applied and new observation observed, a particle filter update according to \eqref{eq:belief_up} is performed, and relevant parts of the tree $\mathcal{T}$ are reused in the next step.

The estimation $\hat{Q}(b,a)$ is used not only for determining the policy $\pi(b)$ but also during the construction of the tree $\mathcal{T}$ to select the actions for propagating particles. 
First, actions that have not been picked already for the belief are chosen randomly with uniform distribution. 
When there are no unused actions left, the {\it upper confidence bound} (UCB) is employed to address the exploration-exploitation trade-off:
\begin{equation}
a_{\mathrm{sel}} = \argmax_{a \in A}{\left(\hat{Q}(b,a) + c\sqrt{\frac{\log{\sum_a|H(b,a)|}}{|H(b,a)|}}\right)},\
\end{equation}
where $c$ is the tunning UCB parameter.

The current belief is updated after the action $\pi(b_0)$ is applied, and new measurements are obtained. The corresponding branches of the belief tree are reused, with the current belief being the new root of the tree. The whole process is repeated until the goal is reached. If needed, new particles are generated to maintain the $n_\mathrm{par}$ number of particles to avoid particle depletion.
\nomenclature{\(\mathcal{Q}\)}{ POMDP -- Q-value}
\nomenclature{\(H\)}{ POMDP -- set of episodes (histories)}
\nomenclature{\(h\)}{ POMDP -- an episode (history)}
\nomenclature{\(h\)}{ POMDP -- UCB parameter}

\subsection{Topology Discretization}
\begin{figure*}[h!]
  \centering
  \hspace{1cm}
   {\epsfig{file = 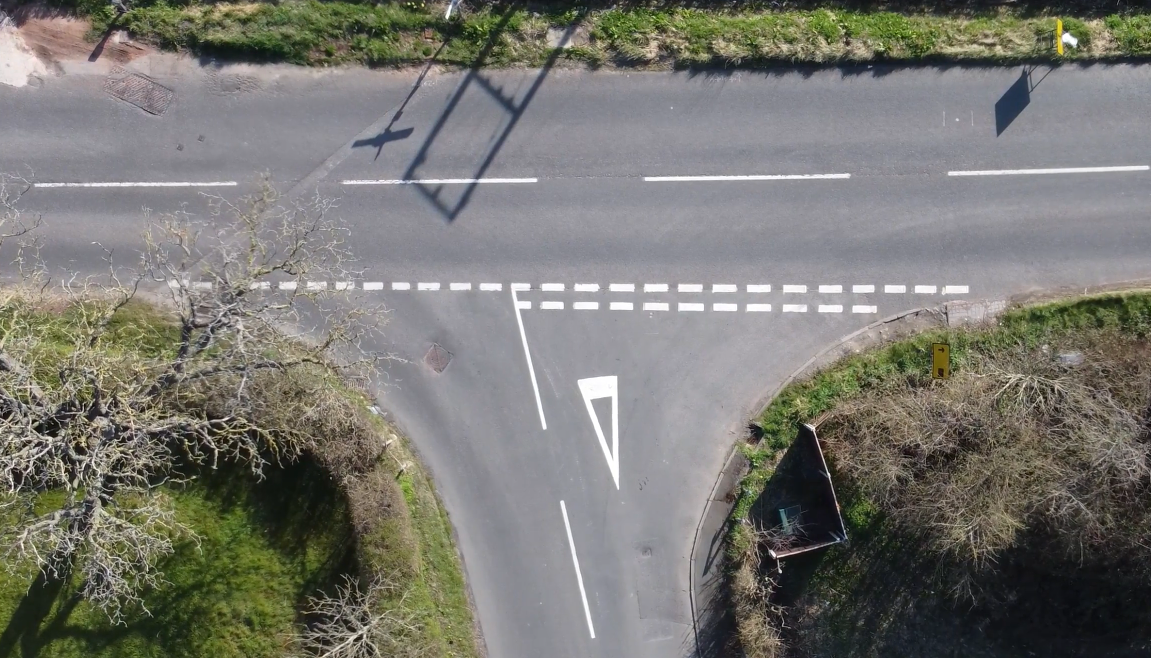, width = 0.9\columnwidth}}
   \hfill 
    {\epsfig{file = 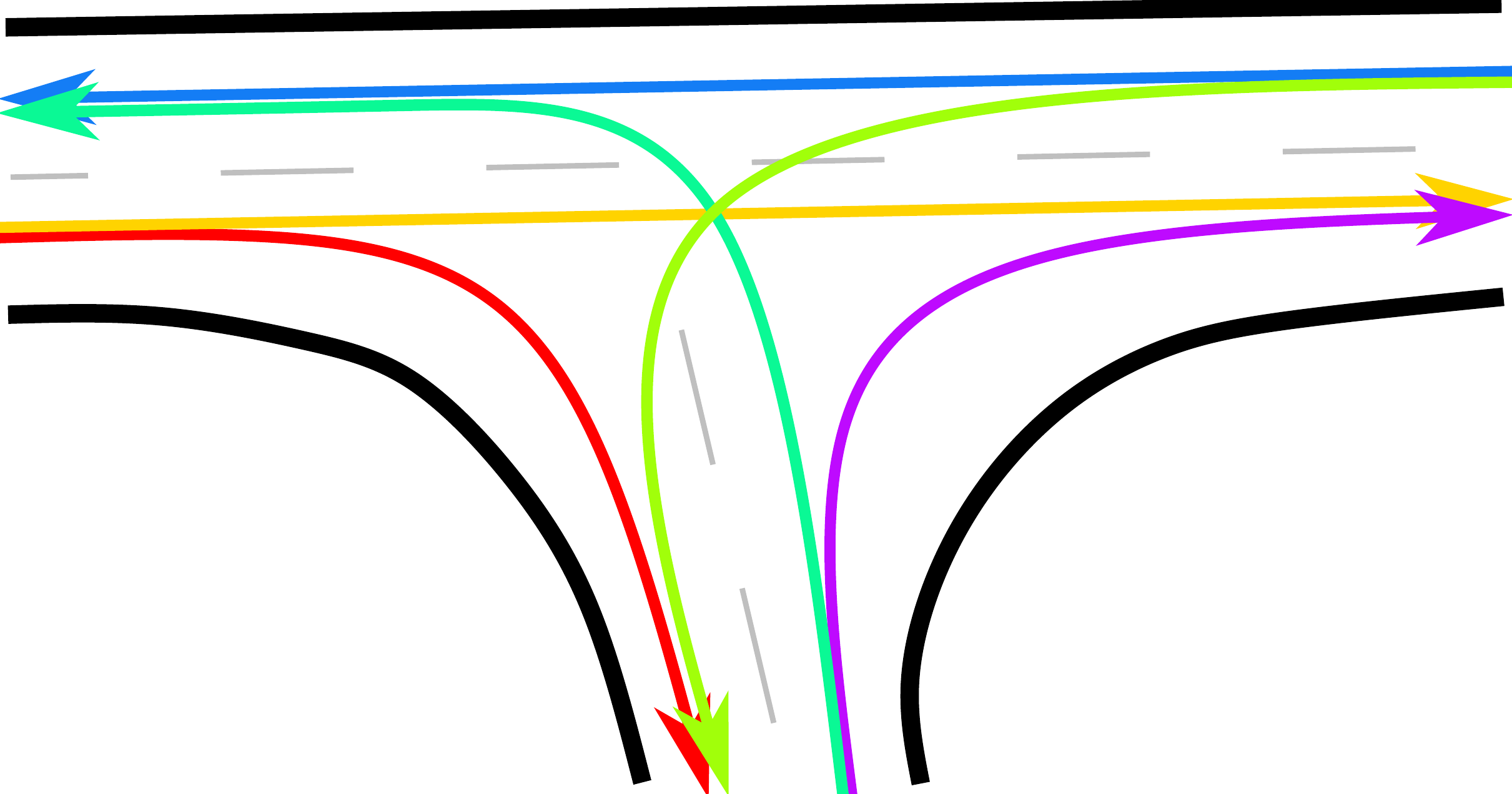, width = 0.9\columnwidth}}
    \hspace{1cm}
  \caption{Visualisation of topology discretization -- an intersection (left) with all possible simple paths (right).}
  \label{fig:topo}
 \end{figure*}

To simplify the representation of the intersection within the POMDP and to reduce computational demands, we adopt an approach from~\cite{hubmannAutomatedDrivingUncertain2018a}. The intersection is condensed into a set of all possible paths, as illustrated in Figure \ref{fig:topo}. Particularly effective for unsignalized intersections, where only a few paths are reasonable to assume, this approach reduces the description of any vehicle to a triplet of values $\bm{s}_i = [p_{i}\,,\ v_{i} \,,\ \mu_{i}]^\top$, where $\mu \in \{1, \dots, m\}$ is the intended path out of the $m$ possible paths, $p$ is the position along the path $p$, and $v$ is the longitudinal velocity. The state of our POMDP is then represented by a vector $\bm{s}$ containing these values for all $n+1$ vehicles, including the ego vehicle.
\label{subsec:topology}
 
To identify possible paths, we utilize the {\it lanelet2} \cite{poggenhansLanelet2HighdefinitionMap2018} map in the OSM XML format as a source for the full intersection topology. Using the lanelet2 library, we generate a route graph, searching for all possible paths as a sequence of $\mathbb{R}^2$ points. Subsequently, each path is stored as two separate cubic splines for the $x$ and $y$ coordinates. This setup provides an efficient mapping from the path position to the global map coordinates, denoted as $\bm{\phi}(p): p \rightarrow \mathbb{R}^2$, and mapping to the unit heading $\bm{\psi}(p): p \rightarrow \bm{y} \in \mathbb{R}^2: \| \bm{y}\| = 1$. For the inverse mapping converting global coordinates back to the position on a path, denoted as $\epsilon(\bm{p}): \bm{p} \rightarrow \mathbb{R}$, we employ a search algorithm to determine the nearest point on the selected path.

\subsection{Dynamics Model}
\label{subsec:models}
As we proceed with the formulation of our POMDP, it is crucial to introduce a dynamical model describing how the state, defined in the previous section, evolves over time. While various mathematical models of different complexities exist for vehicles, we aim for simplicity and reduced computational demand. Thus, we follow the approach from \cite{hubmannAutomatedDrivingUncertain2018a} and adopt a discrete linear point mass model for each vehicle $i \in \{1, \dots, n\}$:
\begin{equation}
\label{eq:model}
\begin{aligned}
\bm{s}^{k+1}_i &= 
\bm{A} \bm{s}^k_i
+ \bm{B} u^k_{i}(\bm{s}) + \bm{\eta}\,,\quad \bm{\eta} \sim \mathcal{N}(\bm{0},\bm{Q})\,,\\
\bm{A} &= \begin{bmatrix} 1 & \Delta  t & 0 \\ 0 & 1 & 0 \\ 0 & 0& 1  \end{bmatrix},\qquad\bm{B} = \begin{bmatrix} \frac{1}{2} {\Delta  t}^2 \\ \Delta  t \\ 0\end{bmatrix} \,.
\end{aligned}
\end{equation}
Here, $\Delta  t$ represents the time between subsequent samples $k$ and $k+1$, and $u(\bm{s})$ denotes the acceleration of the vehicle. The additive Gaussian noise $\bm{\eta}$ with zero mean and covariance matrix $\bm{Q}$ accounts for model inaccuracies.

The first two rows of the model matrices $\bm A$, $\bm B$ represent the change of position $p_i$ and velocity $v_i$ in time. Upon further inspection, it is also noticeable from their last row that the intention $\mu_i$ remains constant. Yet, it is unknown for all vehicles except for the ego. The same applies to acceleration values, which need to be predicted to estimate the future movement of the other vehicles. 

In \cite{hubmannAutomatedDrivingUncertain2018a}, the authors used a heuristically chosen deceleration when a crash is anticipated for other vehicles, otherwise, the reference velocity is followed. Alternatively, a dynamic Bayesian network prediction is presented in the follow-up paper \cite{schulzLearningInteractionAwareProbabilistic2019a}. Another approach \cite{beyPOMDPPlanningRoundabouts2021a} introduces a version of the intelligent driver model (IDM) \cite{treiberCongestedTrafficStates2000}. 

We implemented the IDM-based model for its simplicity:
\begin{equation}
\label{eq:IDM}
u^k_i = a_\textrm{max} \rho_\textrm{IDM} + \omega_1\,,\quad  \omega_1 \sim \mathcal{N}(0,\sigma_{\omega_1}),
\end{equation}
where $a_\textrm{max}$ is the maximal acceleration, and $\omega_1$ introduces randomness into the model with variance $\sigma_{\omega_1}$. The factor $ \rho_\textrm{IDM}$ is given by the IDM:
\begin{equation}
\rho_\textrm{IDM} = 1 - \left( \frac{v_i^k}{v_\textrm{des}}\right)^\delta - \left(\frac{d^{*}(v^k_i,v_\textrm{lead})}{d_\textrm{lead}}\right)^2,
\end{equation}
\begin{equation}
d^{*}(v^k_i,v_\textrm{lead}) = d_\textrm{min} + v^k_i\tau + \frac{v^k_i(v^k_i-v_\textrm{lead})}{2\sqrt{a_\textrm{max} |a_\textrm{min}|}},
\end{equation}
where $v_\textrm{des}$ is the desired velocity, $\tau$ is the time headway, $\delta$ is the acceleration exponent,  $v_\textrm{lead}$ is the velocity of the leading (approached) vehicle, and $a_\textrm{min}$ is the minimal acceleration (maximal deceleration). 

The acceleration of the ego vehicle is the input of our POMDP, obtained from the resulting policy according to~\eqref {eq:policy}. The model~\eqref{eq:model} together with other vehicles acceleration predictions \eqref{eq:IDM} represent the POMDP dynamics model referred to as $T$ in Section \ref{subsec:pomdp}.
\subsection{Observation Model}
\label{subsec:obs}
\begin{figure*}[!b]
  \centering
    \hspace{0.5cm}
   {\epsfig{file = 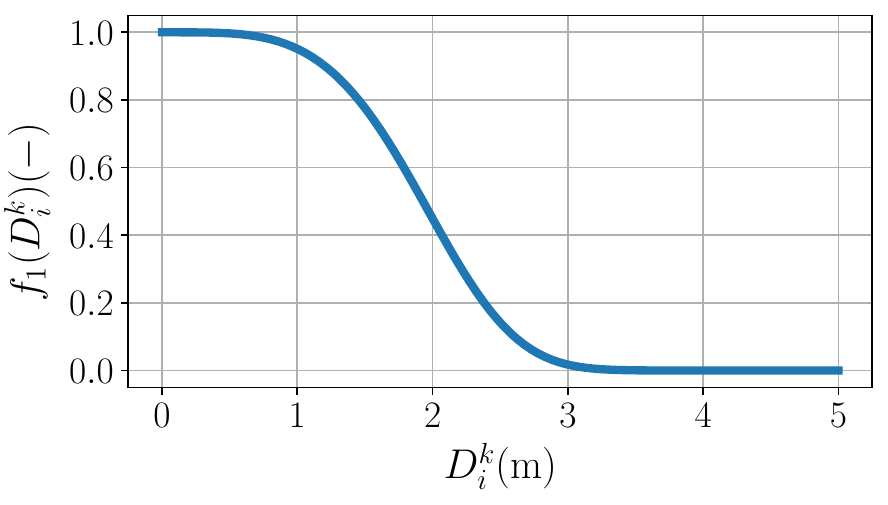, width = 0.9\columnwidth}}
    \hfill 
    {\epsfig{file = 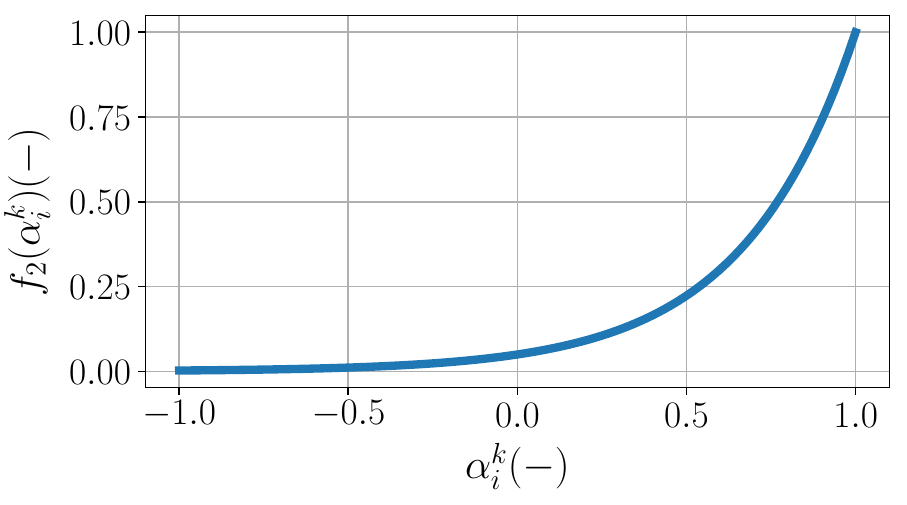, width = 0.9\columnwidth}}
      \hspace{0.5cm}
  \caption{Likelihood functions of features-- $f_1$ (left), $f_2$ (right). }
  \label{fig:example1}
 \end{figure*}
Having established the dynamics model, our focus shifts to the observation model, describing how the observations are obtained from the state of the system. In this context, observations should align with measurements, encompassing variables such as position, velocity, and orientation.

Let vector $\bm{z}_i^k$ denote the observation of road user $i$ at time step $k$. These observations are acquired through spline mappings, as detailed in Section~\ref{subsec:topology}:
\begin{equation}
\label{eq:measurement_model}
\bm{z}_{i}^k = \begin{bmatrix}\bm{\phi}_{\mu_i^k}(p^k_i)\\ {v^k_i} \bm{\psi}_{\mu_i^k}(p^k_i)\\ \bm{\psi}_{\mu_i^k}(p^k_i) \end{bmatrix} + \bm{\zeta},\quad \bm{\zeta} \sim \mathcal{N}(\bm{0}, \bm{R}),
\end{equation}
where $\mu_i^k$ signifies the specific path for the mappings, and $\bm{\zeta}$ represents observation noise following a Gaussian distribution with zero mean and covariance matrix $\bm{R}$.

For particle generation within the particle filter, the estimation of the intended path $\mu_i^k$ from global measurements is essential. As the inverse mapping $\epsilon$ has already been defined, the focus is on estimating the intended path $\mu_i^k$. However, this might not be determined uniquely due to potential path overlaps or shared segments, making paths indistinguishable in certain instances. Consequently, a probability distribution is employed to assess the likelihood of each path intention. This involves combining the likelihoods $f_1$ and $f_2$ of two features shown in Figure~\ref{fig:example1}:
\begin{equation}
    f_1(i,\mu) = e^{\left(-0.05 D^k_i(\mu)^4 + 1)\right)},
\end{equation}
\begin{equation}
    f_2(i,\mu) = e^{\left(3(\alpha_i^k(\mu)-1)\right)}.
\end{equation}
The features are the distance from the closest point projection $D^k_i(\mu) = \|\bm{p}_i^k - \bm{\phi}_{\mu}(\epsilon_{\mu}(\bm{p}_i^k)) \|$, and a dot product $\alpha_i^k(\mu) = \bm{\theta}_i^k \cdot \bm{\psi}_{\mu}(\epsilon_{\mu}(\bm{p}_i^k)) $ of the path heading and measured heading of a vehicle.

The combination of these likelihoods results in the distribution over all path intentions $\mu$:
\begin{equation}
P(i,\mu) = \frac{ q(\mu) f_1(i,\mu)f_2(i,\mu)}{\sum\nolimits_{j=1}^m q(j) f_1(i,j)f_2(i,j)}\,,
\end{equation}
where $ q(\mu)$ is the overlap coefficient, adjusting the probabilities for situations involving overlapping paths, as commonly encountered in roundabouts. 

\subsection{Reward Function}
\label{subsec:reward}
The reward function is a crucial component that defines the behavior of the ego vehicle within the intersection. It is responsible for aligning the vehicle's actions with our desired objectives: efficient navigation through the intersection while prioritizing safety, avoiding excessive acceleration, and preventing collisions. This multi-goal task is mathematically formulated as
\begin{equation}
r^k = r_\mathrm{vel}(\bm{s}_0^k) + r_\mathrm{acc}(a_0^k) + r_\mathrm{crash}(\bm{s}^k)\,.
\end{equation}
The reward at each time step, denoted as $r^k$, is composed of three fundamental terms: $ r_\mathrm{vel}$ representing velocity, $r_\mathrm{acc}$ for acceleration, and $r_\mathrm{crash}$ pertaining to collision avoidance.

The velocity reward encourages the ego vehicle to maintain an appropriate speed throughout its trajectory. It is determined based on the deviation from the reference velocity $ \Delta v = v_\textrm{des} - v^k_0$:
\begin{equation}
r_\mathrm{vel}(v_0^k) =
\begin{cases}
-R_\mathrm{vel} \Delta v &\textrm{if $\Delta v  \geq 1\,,$}\,\\
-R_\mathrm{vel} {\Delta v}^2 &\textrm{otherwise,}
\end{cases}
\end{equation} 
where $R_\mathrm{vel}$ is a positive weight. The desired velocity $v_\textrm{des}$ is influenced by factors such as the curvature $\kappa_\mu(p_0^k)$ at the vehicle's position $p_0^k$ along its intended path $\mu_0^k$. The desired velocity is determined as follows:
\begin{equation}
v_\textrm{des} = \min{\left( v_\textrm{curve}, v_\textrm{lim}\right)},
\end{equation}
where $v_\textrm{curve} = \sqrt{a_{\textrm{lat},\textrm{max}}/\kappa_\mu(p_0^k) }$ is the velocity limited by the maximal acceptable lateral acceleration during cornering, and $v_\textrm{lim}$ is the velocity speed limit of the intersection.

An acceleration penalty is imposed to prevent abrupt changes in velocity, thus increasing driving comfort. It is computed using a quadratic penalty function, weighted by $R_\mathrm{acc} \geq 0$:
\begin{equation}
r_\mathrm{acc}(a_0^k) =
-R_\mathrm{acc} {a_0^k}^2.
\end{equation}

The collision avoidance component is designed to prevent crashes. It assigns a negative reward $-R_\mathrm{crash}$ when a collision is detected and $0$ otherwise:
\begin{equation}
r_\mathrm{crash}(\bm{s}^k) =
\begin{cases}
-R_\mathrm{crash} &\textrm{if ego crashed,}\\
0 &\textrm{otherwise.}
\end{cases}
\end{equation} 
Collision detection relies on the overlap of bounding rectangles of vehicles, as explained in Section~\ref{subsec::formal_def}.
\section{\uppercase{Results}}
\label{sec:results}
\begin{figure*}[!h]
  \centering
   {\epsfig{file = 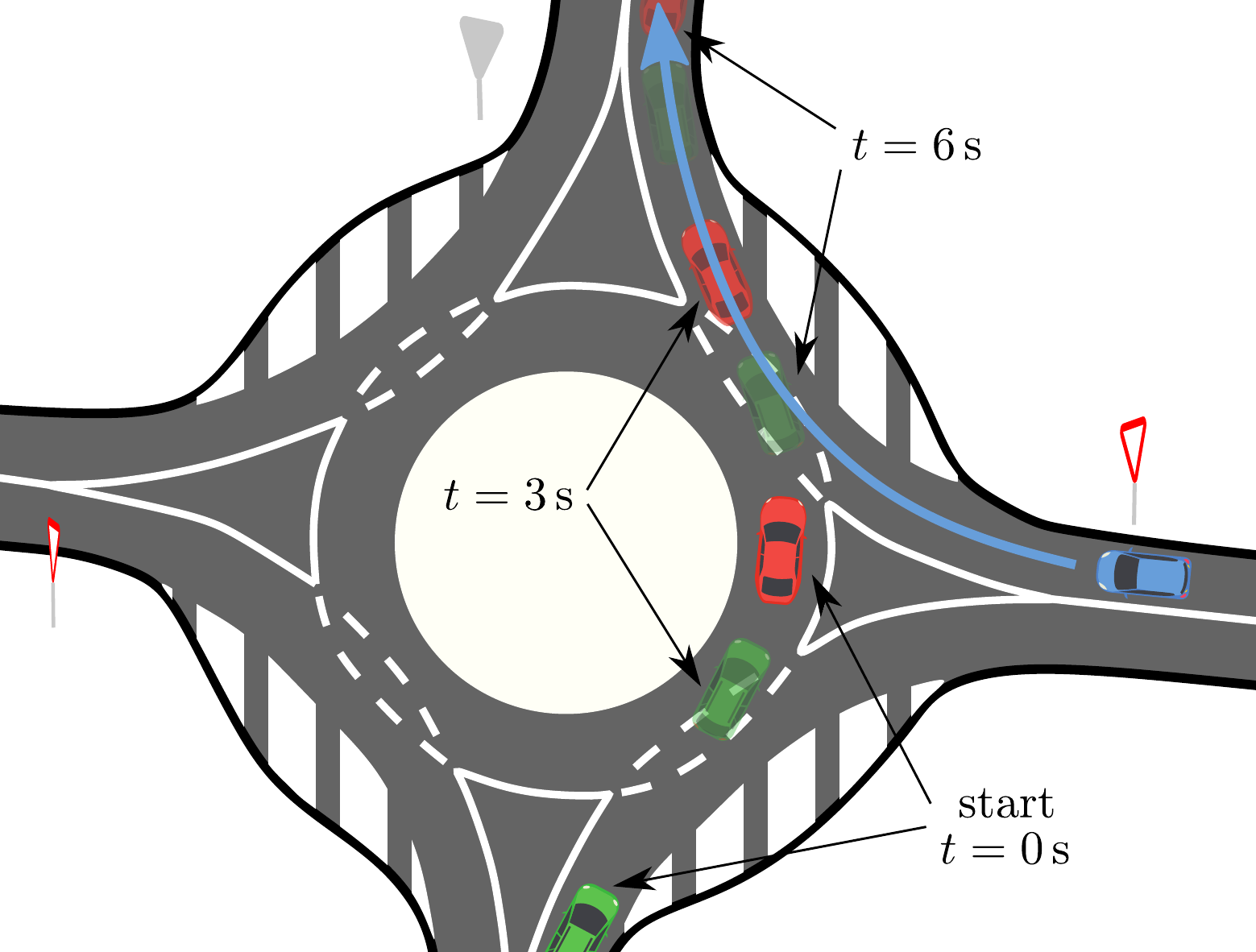, width = 0.95\columnwidth}}
    {\epsfig{file = 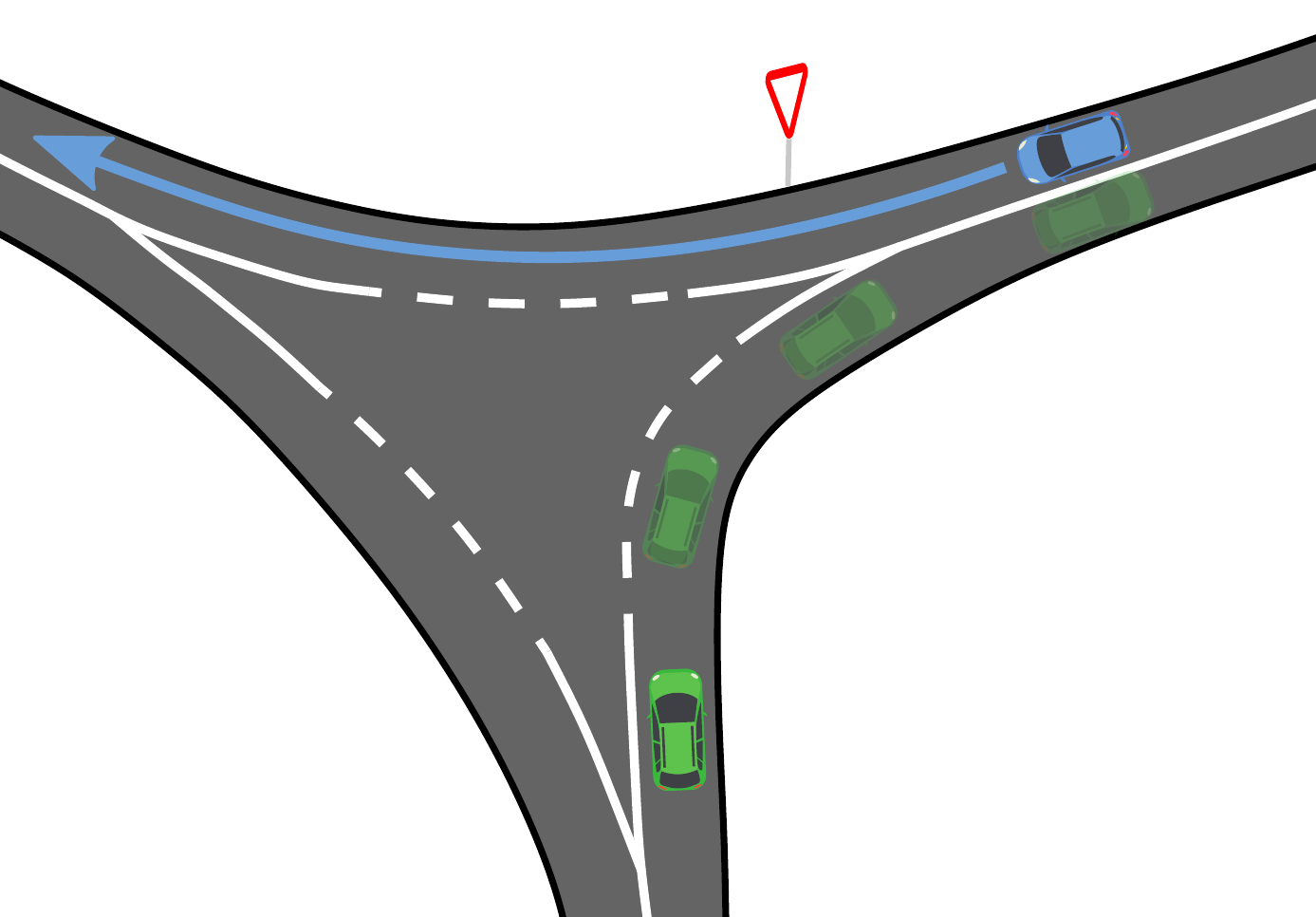, width = 0.95\columnwidth}}
  \caption{Simulation scenarios -- roundabout (left), threeway junction (right). The ego car is represented with a blue color.}
  \label{fig:scenarios}
 \end{figure*}

In this section, we test the investigated method's collision-free planning capabilities and the impact of parameter adjustments on the planning algorithm's effectiveness. For these proposes, we ran simulations based on data collected from aerial observations of two specific intersections as shown in Figure~\ref{fig:scenarios}. In the simulations, building upon the results of~\cite{vsimon2021simulavcni} we replaced a yielding vehicle with our ego vehicle to mimic real-life situations. 
Even though the use of such offline data does not include direct reactions of other vehicles to the ego vehicle, we deem it appropriate for testing situations where the ego is expected to yield and adequately respond to the behavior of others. Also, our simulations assume that the ego vehicle adheres to the point mass model~\eqref{eq:model}. 
 
 We focus on the Adaptive Belief Tree (ABT) algorithm's parameters: the optimization horizon $N$, the number of particles $n_\mathrm{par}$, the number of episodes $n_\mathrm{ep}$, and the Upper Confidence Bound (UCB) parameter $c$ influencing the exploration-exploitation trade-off. These parameters are crucial to the ABT algorithm, and their individual impacts on the performance are not immediately apparent. The default values of all parameters used in our simulations are provided in Table \ref{table:parameters}. We obtained those values by tuning or by following the works they were used in.

\begin{table}[t]
\caption{Default parameter values}
\label{table:parameters}
\centering
\begin{tabular}{|c|c|c|}
\hline
 & Description & Value \\ \hline
$A$            & action set      & $\left\{-2,-1,0,1 \right\}$ \si{\meter \per \square \second}                 \\ \hline
$c$            & UCB parameter        & 20000                  \\ \hline
$N$             & opt. horizon        & 5                 \\\hline
$n_\mathrm{ep}$             & num. of episodes       & 3000                  \\\hline
$n_\mathrm{par}$              & num. of particles       & 300                  \\\hline
$\gamma$              & discount fact.       & 1                 \\\hline
$\bm{Q}$            & dyn. covariance       & $\bm{0} $                 \\ \hline
$\bm{R}$            & obs. covariance       & $\mathrm{diag}(10^{-2},10^{-2},0)$                 \\ \hline
$\sigma_{\omega_1}$            & IDM st. dev.      & 1       \si{\meter \per \square \second}           \\ \hline
$\tau$              & time headway       & 2 \si{\second}               \\\hline
$\delta$              & acc. exponent       & 4              \\\hline
$d_\mathrm{min}$              & IDM min. dist.       & 1 \si{\meter}               \\\hline
$R_\mathrm{acc}$            & acc. reward coef.      & 1                  \\ \hline
$R_\mathrm{vel}$            & vel. reward coef.       & 100                  \\ \hline
$R_\mathrm{crash}$              & crash reward        & 10000                 \\ 
\hline
$a_{\textrm{lat},\textrm{max}}$              & max. lat. acc.     & 0.5  \si{\meter \per \square \second}   \\ \hline

\end{tabular}
\end{table}

For each scenario, we tested every parameter combination in 50 simulations,  employing three different time step lengths $\Delta t$. The chosen parameter settings significantly influence the simulation runtime, except for the UCB factor $c$, which does not alter the runtime. To provide a clearer perspective on how these settings impact the simulation duration, Table \ref{tab:runtime} presents a comparison of roundabout scenario runtime estimates for two distinct parameter configurations alongside their corresponding real-time percentages. Our simulation software was implemented in C++, utilizing the TAPIR toolbox together with ROS~1, and we ran the simulations on a laptop with an Intel i5-11500H~CPU.

\begin{table}[t]
\caption{Runtime estimate with realtime percentages}
\begin{tabular}{|c|c|c|}
\hline
 $\Delta t$ (s)&\makecell{ Runtime (s) \\ default settings} &\makecell{ Runtime (s) \\  $n_\mathrm{ep} = 10^4$, $n_\mathrm{par} = 2\cdot 10^3$}\\ \hline
 1& 6 (75\%) & 14 (175\%)\\ \hline
 0.5& 10 (125\%)& 42 (525\%) \\ \hline
 0.25& 24 (300\%)& 118 (2350\%)\\ \hline 
\end{tabular}
\label{tab:runtime}
\end{table}
\subsection{Roundabout Scenario}
\label{subsubsec:roundabout_sim}
The first scenario involves the ego vehicle approaching a roundabout at 8 \unit{\meter \per \second}, interacting with two other vehicles, as seen in the left image of Figure \ref{fig:scenarios}. This scenario tests the vehicle's ability to yield and avoid crashes.

\begin{figure*}[t]
  \centering
   {\epsfig{file = 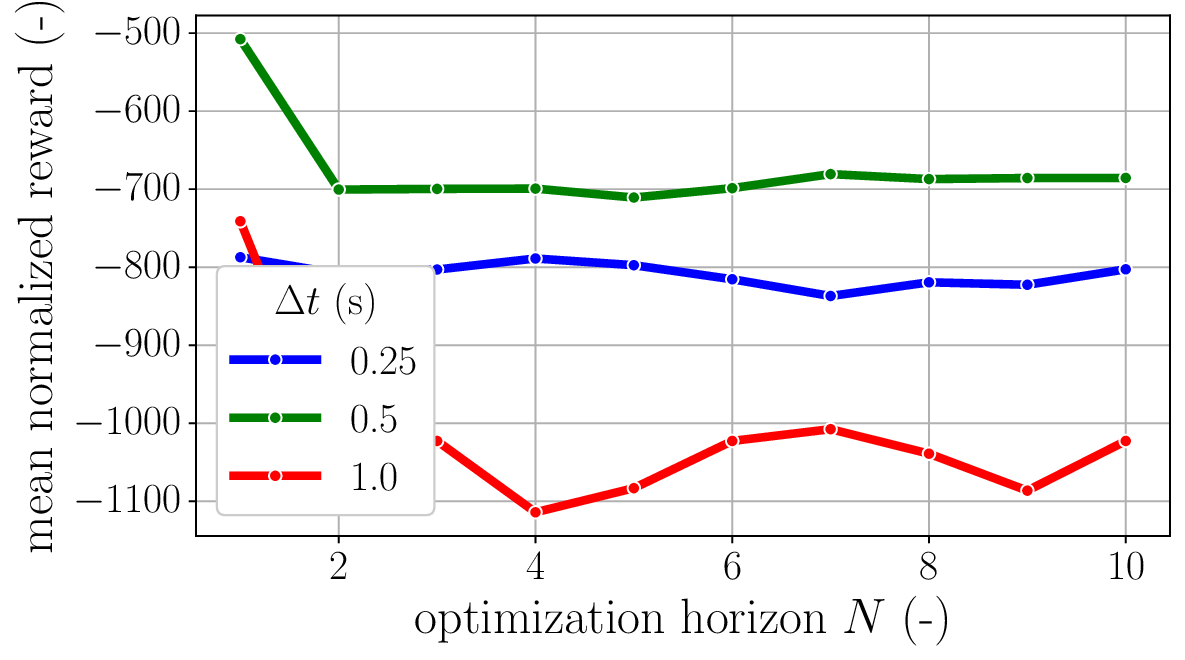, width = 1\columnwidth}}
{\epsfig{file = 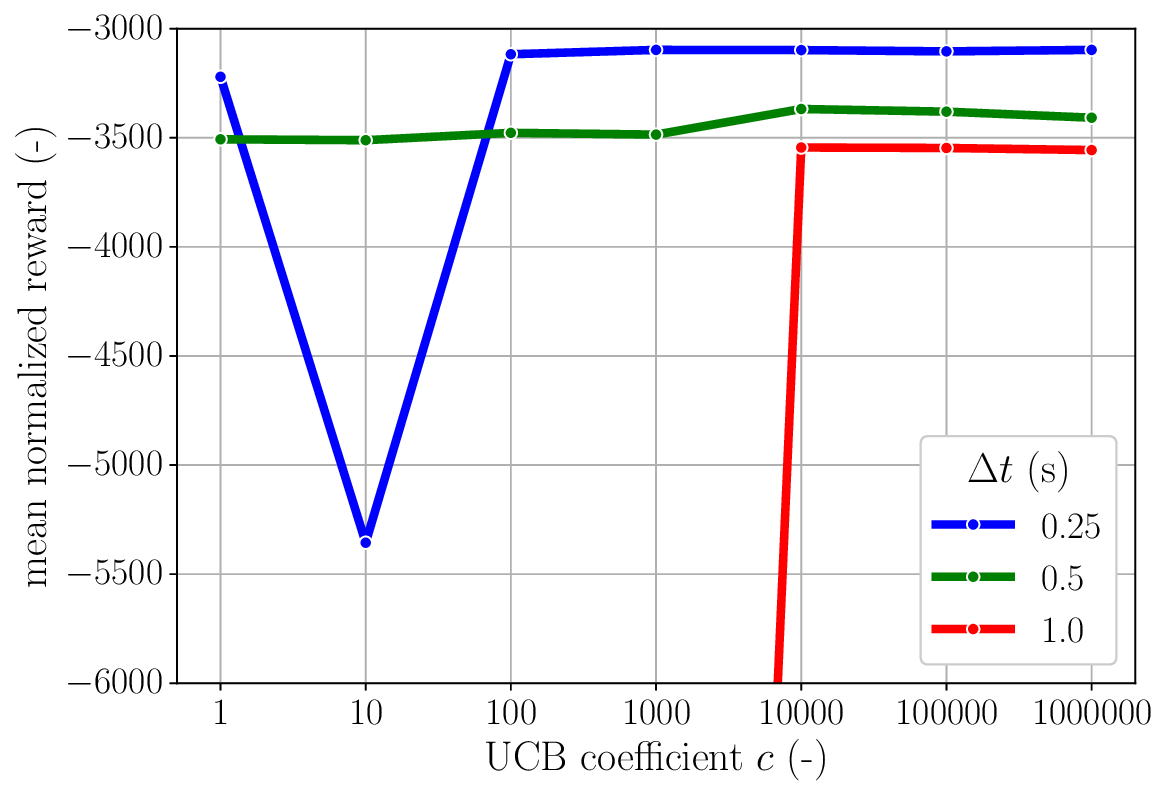, width = 1\columnwidth}}
  \caption{Roundabout rewards -- varying optimization  horizon $N$ (left), and varying UCB coefficient  $c$ (right). }
  \label{fig:roundabout_sim1}
 \end{figure*}

\begin{figure}[!h]
  \centering
{\epsfig{file = 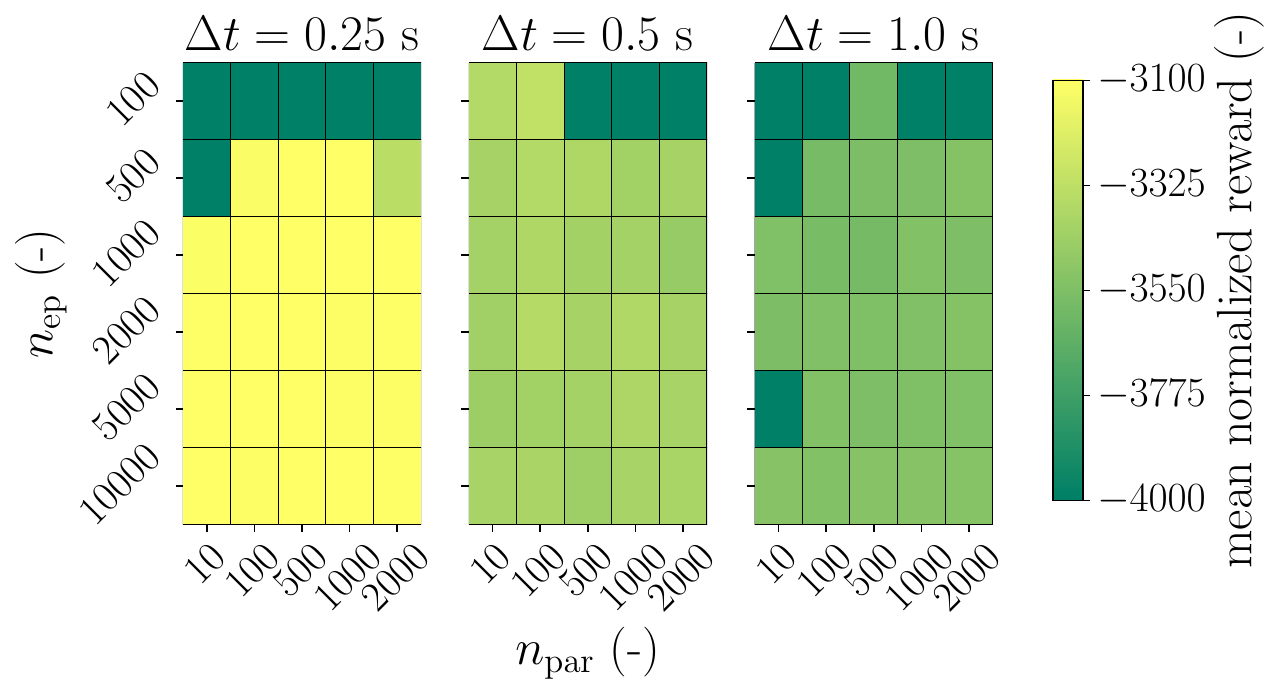, width = 1\columnwidth}}
  \caption{Roundabout rewards for varying $n_\mathrm{par}$ and $n_\mathrm{ep}$.}
  \label{fig:roundabout_sim2}
 \end{figure}

The results, depicted in Figures~\ref{fig:roundabout_sim1} and~\ref{fig:roundabout_sim2}, show the mean of normalized reward $\Delta t \sum_i{r^i}$ achieved for various settings. In most cases, the simulation resulted in no crashes with reward values between $-3200$ and $3600$ based on $\Delta t$. The smaller the time step, the higher the reward because the planning algorithm has more samples to react.

Our key findings include identifying parameter threshold values, where their further changes have minimal impact on performance yet significantly increase computation time.
The threshold for $N$ is 2; since for $N=1$, vehicles frequently crash at sample times $\Delta t \leq0.5$ \unit{\second}.
For $c$, the threshold appears to be $10000$, with regular crashes occurring at lower values, mainly for $\Delta t = 1$ \unit{s}. Crashes also happened commonly for $n_\mathrm{par} \leq 100$ and for $n_\mathrm{ep} \leq 500$.

\subsection{Threeway Junction Scenario}
\label{subsubsec:threewa_sim}
In the second scenario, the ego vehicle approaches a three-way junction at 8 \unit{\meter \per \second}, while another vehicle with the right of way also approaches the intersection as depicted in the right image of Figure \ref{fig:scenarios}. There is no danger of crashing because the other vehicle turns in a non-collision direction, yet this is an unknown behavior initially. This scenario examines the ego vehicle's information-gathering capability.

Note that the reward here does not indicate the total quality of the solution; rather, it says how well the vehicle follows the reference velocity. Slowing down to gather more information is reflected as a lower~reward. 

\begin{figure*}[t]
  \centering
   {\epsfig{file = 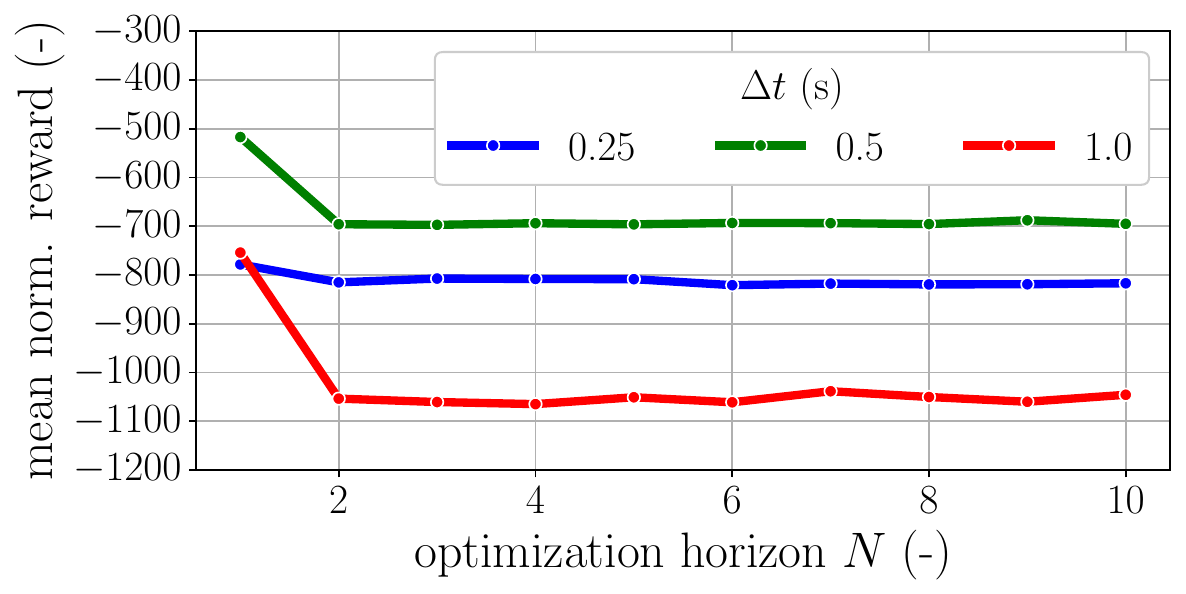, width = 1\columnwidth}}
{\epsfig{file = 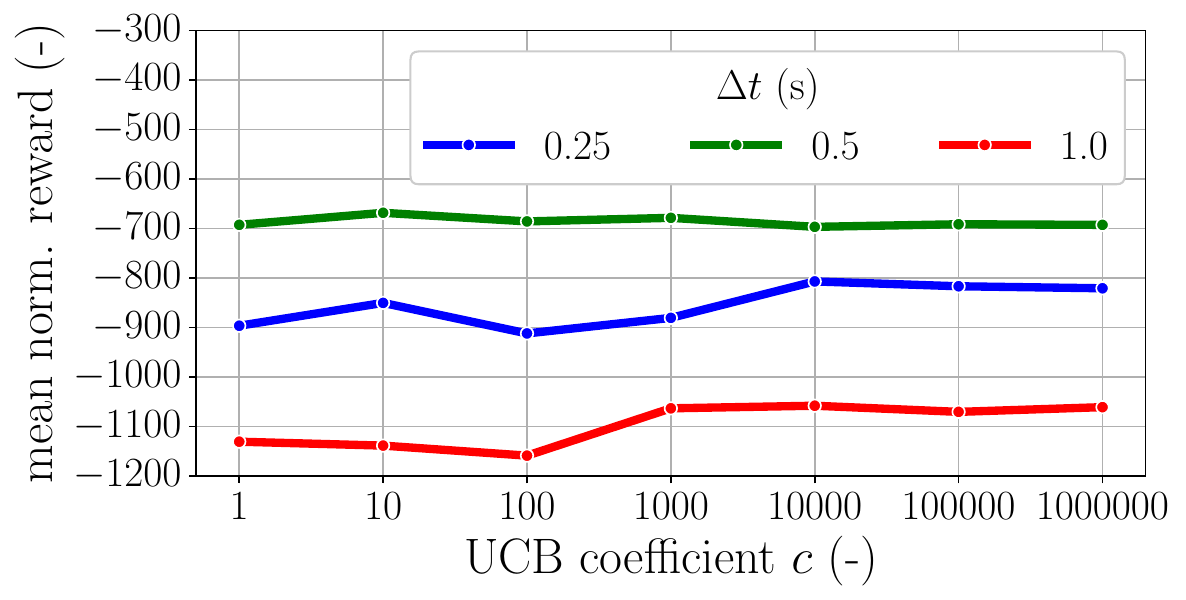, width = 1\columnwidth}}
  \caption{Threeway rewards -- varying optimization  horizon $N$ (left), and varying UCB coefficient  $c$ (right). }
  \label{fig:threeway_sim1}
 \end{figure*}

\begin{figure}[t]
  \centering
{\epsfig{file = 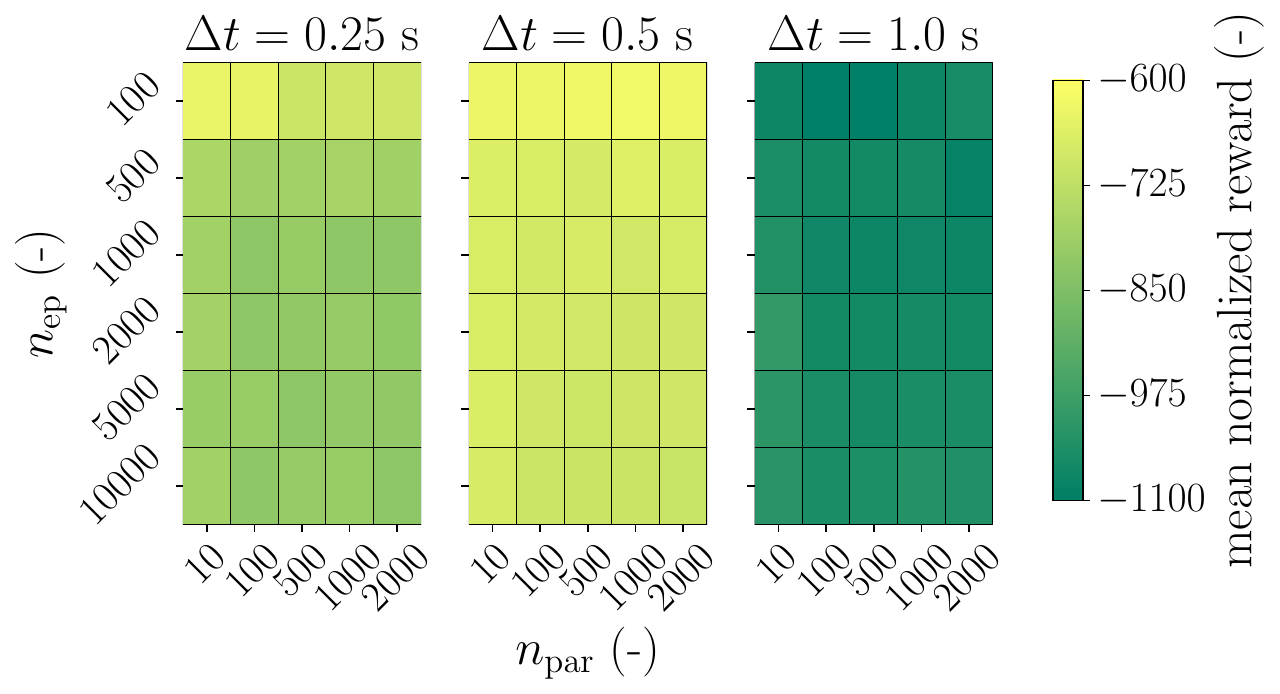, width = 1\columnwidth}}
  \caption{Threeway rewards for varying $n_\mathrm{par}$ and $n_\mathrm{ep}$.}
  \label{fig:threeway_sim2}
 \end{figure}

This is reflected in our results shown in Figures~\ref{fig:threeway_sim1}, and \ref{fig:threeway_sim2}, where parameter combinations of $N$, $n_\mathrm{par}$, and $n_\mathrm{ep}$ that caused crashes in the roundabout scenario yield higher rewards as they ignore the possible danger of crashing. The reference velocity is not followed properly for the $c$ values below the threshold, making the rewards lower, especially for $\Delta t = 1$~\unit{s}, and $\Delta t = 0.5$~\unit{s}. Also, the rewards for $\Delta t = 0.25$~\unit{\second} are generally lower than for $\Delta t = 0.5$~\unit{\second}, indicating that the cost of delaying a decision when more samples are available is lower relative to the total reward. This behavior might be fine-tuned~by~parameters~$R_\mathrm{acc}$,~ $R_\mathrm{crash}$~and~$R_\mathrm{vel}$. 

 
\section{\uppercase{Conclusion}}
\label{sec:conclusion}
In this paper, we successfully verified the planning capabilities of the investigated trajectory planning POMDP-based method for unsignalized intersection crossing. We executed a series of simulations based on real-life data from aerial recordings of two distinct intersections. The method proved capable of handling the uncertain aspects of the problem, such as the unknown intention of other vehicles. 

Additionally, we investigated the influence of parameter adjustment on the method's performance. Our results indicate that there are certain well-performing threshold values. Exceeding them yields little to no gains and increased computation time. This finding is beneficial with regard to the future application of this method and its derivations.


\bibliographystyle{apalike}
{\small
\bibliography{lib}}

\end{document}